\documentclass{llncs}
\usepackage{eccv}
\usepackage{eccvabbrv}
\usepackage{graphicx}
\usepackage{booktabs}
\usepackage{float}
\usepackage[accsupp]{axessibility} 
\usepackage[pagebackref,breaklinks,colorlinks,citecolor=eccvblue]{hyperref}

\usepackage{orcidlink}

\begin{document}

\title{Physics Informed Human Posture Estimation Based on 3D Landmarks from Monocular RGB-Videos} 

\titlerunning{Enhanced Human Posture Estimation}

\title{Physics Informed Human Posture Estimation Based on 3D Landmarks from Monocular RGB-Videos}

\titlerunning{Physics Informed Human Posture Estimation}

\author{
Tobias Leuthold\inst{1} \and
Michele Xiloyannis\inst{1} \and
Yves Zimmermann\inst{2}
}

\authorrunning{Leuthold, Xiloyannis, Zimmermann}

\institute{
Akina AG, Zurich, Switzerland \\
\email{leutholdtobias@gmail.com}, \email{michele.xiloyannis@akina.health}
\and
Interactive Robotics and Health Technology Group, OST – Ostschweizer Fachhochschule, Rapperswil-Jona, Switzerland \\
\email{yves.zimmermann@ost.ch}
}

\maketitle

\begin{abstract}
Applications providing automated coaching for physical training are increasing in popularity, e.g., physical therapy. These applications rely on accurate and robust pose estimation using monocular video streams.
State-of-the-art models like BlazePose excel in real-time pose tracking. However, their lack of anatomical constraints indicates improvement potential by including physical knowledge. 

We present a real-time post-processing algorithm fusing the strengths of BlazePose's 3D and 2D estimations using a weighted optimization, penalizing deviations from expected bone length and biomechanical models. Bone length estimations are refined to the individual anatomy using a Kalman filter with adapting measurement trust.

Evaluation using the Physio2.2M dataset shows a 10.2\% reduction in 3D MPJPE and a 16.6\% decrease in errors of angles between body segments compared to BlazePose's 3D estimation. Our method provides a robust, anatomically consistent pose estimation based on a computationally efficient video-to-3D pose estimation, suitable for automated physiotherapy, healthcare, and sports coaching using consumer-level laptops and mobile devices. The refinement runs on the backend with anonymized data only.

\end{abstract}

\section{Introduction}
\label{sec:intro}

\begin{figure}[H]
    \centering
    \includegraphics[width=1\textwidth]{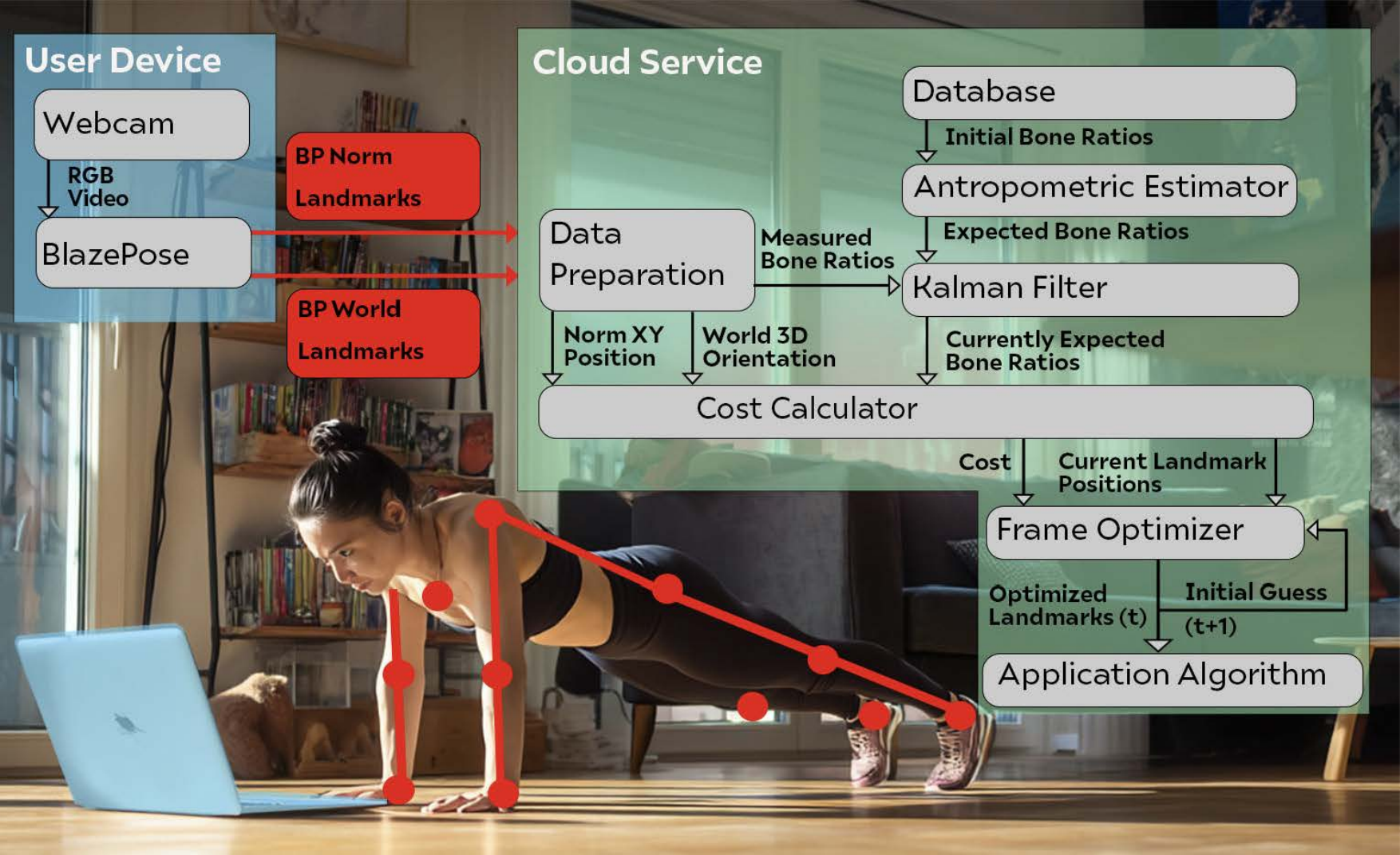}
    \caption{Pipeline Overview. The arrows represent information flow, while the boxes represent modules in the pipeline responsible for processing inputs and generating outputs. Blue is used for the parts that run on the user device, while green represents the parts that run on our backend. All video data stays on the user's device.}
    \label{fig:catchy_img}
\end{figure}

\subsection{Motivation}
Rising healthcare costs \cite{bfs_health_expenses} and limited access to physical therapy highlight the need for scalable, cost-effective solutions such as automated therapy coaching \cite{e-physio-cost-effective, e-physical-activity-effective}. Feedback is a substantial component of movement therapy coaching \cite{feedback-important}. Therefore, computer vision solutions for human posture estimation are a key component of the overall solution. 2D posture estimations enable sufficient movement assessment for some exercises. However, more complex exercises require perception of the 3D posture to detect correct execution and identify mistakes. Further, 3D posture assessments allow for using more convenient orientations of the user towards the sensor \cite{3d-accurate-than-2d}. 

Automated therapy coaching should be accessible with consumer-grade hardware and user-friendly setups. Thus, the posture estimation should work with monocular video streams from any consumer-grade laptop.

\subsection{Related Work}

Monocular, markerless human pose estimation is increasingly favored due to scalability, cost-effectiveness, and ease of deployment in real-world environments, e.g. healthcare applications \cite{suo2024motion, scott2022healthcare}. Unlike traditional multi-camera marker-based or depth-based motion capture systems \cite{rahul2018mocap, colyer2018review, zhang2012microsoft}, monocular systems offer simplified setups balancing accuracy and computational efficiency \cite{avogaro2023markerless}.

State-of-the-art monocular pose estimators like BlazePose \cite{blazepose-pose-estim} excel in computational efficiency, directly estimating 3D poses suitable for real-time use on mobile and consumer hardware \cite{fastape2024}. Alternative approaches commonly employ a two-step method, first estimating 2D keypoints and subsequently lifting them into 3D poses. Recent transformer-based architectures such as MotionBERT \cite{zhu2023motionbert}, PoseFormerV2 \cite{poseformerv2}, and MotionAG Former \cite{mehraban2024motionagformer} have notably improved 3D accuracy by leveraging temporal dynamics but rely heavily on high-quality 2D keypoint detectors like RTMPose \cite{rtmpose-pose-estim} or HRNet \cite{hrnet-pose-estim}.

Incorporating explicit anatomical and biomechanical constraints into pose estimation has been shown to significantly enhance accuracy and anatomical plausibility \cite{chen2021anatomy, cao2020anatomy}. For example, anatomical knowledge such as consistent bone lengths can refine estimations and reduce temporal inconsistencies \cite{chen2021anatomy}. Weakly supervised approaches have also leveraged biomechanical models to constrain predictions, reducing the need for extensive labeled datasets \cite{spurr2020weakly, dabral2018learning}.

Although BlazePose represents the best real-time balance between computational efficiency and accuracy~\cite{fastape2024}, it does not achieve the clinically required accuracy threshold of 12° for physiotherapy applications~\cite{physio12degs}. Like most large neural networks, BlazePose lacks explicit anthropometric constraints, leading to inconsistent bone length estimations across frames. However, BlazePose provides precise spatial orientation through its world landmarks and accurate pixel-space alignment via normalized coordinates~\cite{fastape2024}.

\subsection{Proposed Solution}
We propose an online post-processing algorithm to refine BlazePose's normalized and world landmark estimations by leveraging the strength of both signals. Further, the signal fusion shall integrate anthropometric and biomechanical knowledge to enhance the consistency of the estimation, running the additional methods on the backend without access to video.

Further, we estimate a model of constant relative bone length, promoting anatomical consistency of the pose estimation. We also incorporate biomechanical knowledge to correct the estimation by utilizing the scapulohumeral rhythm, which describes translation of the glenohumeral joint. A key aspect of the proposed solution is sensor fusion, where both the normalized 2D landmarks and the 3D world landmarks from BlazePose are combined. Our method utilizes the accuracy of the normalized landmark estimation in 2D by trusting the corresponding direction of the landmarks w.r.t. the focus point. Depth information is reconstructed by optimizing for consistency with the relative posture provided by the 3D estimation of the world landmarks.  This fusion enhances robustness and accuracy. This combined approach aims to achieve physiotherapist-level accuracy, targeting body angle estimation errors within 12° \cite{physio12degs}, and specifically addresses BlazePose's inconsistent bone length issue, thus improving reliability for clinical use.

\section{Methods}
\label{chap:methods}

The pipeline outlined in Figure \ref{fig:catchy_img} operates in a sequential manner, processing data through several stages to refine landmark positions and ensure robust estimations:

\textbf{Input:} The pipeline starts with an RGB video, which is processed by BlazePose to generate normalized and world landmarks. These landmarks then serve as input to the optimization pipeline.

\textbf{Optimization of Landmarks:} The pipeline minimizes a cost function combining multiple components via the Cost Calculator, which consists of the costs detailed in section \ref{subsec:costs_for_estimation}.

\textbf{Output:} The optimized 3D landmarks are computed for each frame. The optimized landmarks for frame \(t\) are then used as an initial guess for the optimization of the next frame (\(t+1\)).

\textbf{Bone Ratio Adjustment via Kalman Filter:} In the Bone Length Estimator, atomical constraints are applied using a Kalman Filter, which iteratively updates the relative length of the bones to each other (bone length ratios) based on confidence-weighted measurements. The Kalman Filter is initialized with bone length ratios derived from anthropometric data. The estimation of bone length ratios is sufficient, as the posture estimation does not require knowledge about the absolute length.

\subsection{Pipeline Inputs}
\label{sec:pipeline_inputs}

\subsubsection{Output from BlazePose}
\label{sec:blazepose_output}

The BlazePose outputs are provided in two different coordinate systems with different underlying estimation models: 

The normalized landmarks' coordinates are scaled relative to the image dimensions, with the top-left corner at \( (0,0) \) and the bottom-right at \( (1,1) \). The \(z\)-coordinate extends along the camera’s line of sight but lacks a consistent scale, making it unreliable, compared to the accuracy of the \(x,y\)-coordinates. Additionally, the \(z\)-axis is not orthogonal to the \(x,y\)-plane, which complicates its use. Due to these inconsistencies, the normalized \(z\)-coordinate is not used in this study.

World landmarks are provided in Euclidean coordinates with the origin at the midpoint of the hips, measured in meters. The system forms a true orthogonal coordinate system, with the \(z\)-axis perpendicular to the image plane, making it a more reliable choice for posture analysis.

BlazePose also provides confidence metrics with its estimates: a \textit{visibility score}, indicating whether a landmark is within the frame and unobstructed, and a \textit{presence score}, reflecting the probability of a landmark lying within the frame, regardless of occlusion. Both range from 0.0 to 1.0, with higher values indicating greater confidence.

\subsubsection{Initial Bone Ratios}
\label{subsec:initial_bone_ratios}
To promote and accelerate conversion to anatomically plausible solutions, the initial values for the bone ratios were derived from videos of eleven individuals standing in a neutral, upright posture with arms relaxed at the sides and palms facing inward. This approach was preferred over anthropometric datasets like ANSUR2, as BlazePose defines landmarks differently than the anatomical landmarks used in ANSUR2. For instance, ANSUR2 measures pelvis width at the outer sides, whereas BlazePose places hip points more narrowly. Similarly, femur length differs due to variation in knee and hip landmark placement. By using reference videos aligned with BlazePose’s landmark definitions, the initial values for the bone ratios remain consistent with the estimation framework.

\subsection{Data Preparation}
\label{sec:data_preparation}

\subsubsection{Data Filtering}  
\label{subsec:data_filtering}  

To reduce high-frequency noise while preserving motion details, a 4th-order Butterworth low-pass filter with a cutoff frequency of 10 Hz is applied to the \(x, y, z\) world and normalized coordinates at a sampling rate of 30 Hz. The Butterworth filter ensures a smooth signal with minimal delay, making it well-suited for real-time applications. Although lower-delay filters like OneEuro exist, the Butterworth filter maintains consistent delay across signals, preventing distortions when fusing multiple inputs.

\subsubsection{Line of Sight Vectors}
\label{subsubsec:line_of_sight_vectors}
The line of sight vectors represent the lines connecting the camera's focal point to points on the image sensor. Accurate 2D estimates of BlazePose's normalized landmarks imply that corresponding 3D points must lie along vectors. This constraint is incorporated into the cost function (\ref{subsec:costs_for_estimation}).

\subsection{Bone Ratio Adjustment}
\label{subsec:boneratioadjustment}
\subsubsection{Bone Ratio Calculation and Adjustments}
All images are defined in a right‐handed Cartesian frame where the image plane coincides with the \(xy\)-plane and depth extends along the \(z\)-axis.
Limb lengths are estimated in the \(xy\)-plane and corrected for limb orientation. These lengths are then normalized by the summed fixed bone length, which consists of the pelvis, ulna, humerus, femur, and tibia. Ratios exceeding a 15\% deviation from initial values are rejected as outliers, and measurements exceeding \(50^{\circ}\) inclination to the \(xy\)-plane are not used to update the ratios due to the bad signal noise ratio.

The Kalman filter iteratively refines bone ratios, weighing prior estimates and new observations based on confidence scores. Updates occur only if the confidence is above a threshold, and for symmetric limbs, a weighted average of both sides' measurement is used.

\subsubsection{Torso Segment Lengths Depending on Humerothoracic Elevation}
\label{subsec:humerothoracic-elevation-functions}

We define the spine length as the distance between the hip and the midpoint between both glenohumeral landmarks, the lateral trunk length as the distance between the hip and glenohumeral landmarks and the shoulder width as the distance between the glenohumeral landmarks. On humans, these segment lengths are depending on the scapular rotation. The scapular rotation can be estimated by the humerothoracic elevation and typical physiological joint coordination, e.g. scapulohumeral rhythm. Prior research exists on this motion relationship \cite{georgarakis2022supporting}, but due to BlazePose’s unique joint definitions, our own dataset was recorded to ensure compatibility. 

Using the recorded dataset, we create polynomial functions to model the relative changes in shoulder width, lateral trunk length, and spine length in relation to the ipsilateral arm elevation angles. The total spine length adjustment is defined as the average of both arm's individual spine length adjustment. The lateral trunk length is adjusted individually for each side.

\subsubsection{Confidence Score Calculation}
A confidence score quantifies the reliability of limb length estimations for an image frame. It is computed using the landmark's visibility and the limb’s angle relative to the \(xy\)-plane. Higher visibility and a lower inclination result in a higher score. Confidence scores guide the Kalman filter updates, prioritizing high-confidence frames.

\subsection{Posture Estimation}

\subsubsection{Cost Function}
\label{subsec:costs_for_estimation}

The total cost for each frame is defined as:

\begin{equation}
{J}(\mathbf{X}, {}^{\mathcal{W}\!}\mathbf{X},  {}^{\mathcal{N}\!}\mathbf{X}, \mathbf{B} ) = w_W {J}_{W}(\mathbf{X}, {}^{\mathcal{W}\!}\mathbf{X}) 
+ w_S J_S(\mathbf{X}, {}^{\mathcal{N}\!}\mathbf{X})
+ w_B J_B(\mathbf{X}, \mathbf{B} ) 
+ w_M J_M(\mathbf{X}, \mathbf{B})
\end{equation}

where \( w_W, w_S, w_B, w_M \) are the weights assigned to each cost component, controlling their relative influence on the total cost. Superscripts \(\mathcal{N}\) and \(\mathcal{W}\) denote variables expressed in normalized and world coordinates, respectively. 

Specifically, \(\mathbf{X}\) represents the landmark coordinates, optimized during the process. \(\mathbf{B}\) stands for the bone model which defines the bone ratios with respect to each other.

\textbf{World Likeliness Cost (\(J_W\))}: Ensures alignment between optimized coordinates and BlazePose world landmarks. It consists of two components: a \textit{relative orientation cost} (\(J_{W}^{rel}\)), penalizing deviations in angles between limb segments, and an \textit{absolute orientation cost} (\(J_{W}^{abs}\)), ensuring limbs maintain their overall orientation.

\begin{equation}
{J}_{W}(\mathbf{X}, {}^{\mathcal{W}\!}\mathbf{X}) = \sum_{i \in l}^NJ_{W,i}^{rel}(\mathbf{X}, {}^{\mathcal{W}\!}\mathbf{X}) + J_{W,i}^{abs}(\mathbf{X}, {}^{\mathcal{W}\!}\mathbf{X})
\label{eq:angle_cost_total}
\end{equation}

Where \(N\) is the number of limbs and \(l\) denotes the set of limbs.

The relative world likeliness cost is computed as:
\begin{equation}
J_{W,i}^{rel}(\mathbf{X}, {}^{\mathcal{W}\!}\mathbf{X}) = \sum_{j \neq i}^N\left({{}^{\mathcal{W}\!}\alpha_{ij}({}^{\mathcal{W}\!}\mathbf{X}) - \alpha_{ij}}(\mathbf{X})\right)^2{\lambda_1}
\label{eq:J_W_multi}
\end{equation}
where \(\alpha_{ij}\) and \({}^{\mathcal{W}\!}\alpha_{ij}\) denote the angles between limbs \(i,j\) in optimized and BalzePose world coordinates, respectively.

The absolute orientation cost is defined by the normalized scalar product \(\xi\) between optimized limb vectors \(\mathbf{k}_i\) and their world counterparts \({}^{\mathcal{W}\!}\mathbf{k}_i\):

\begin{equation}
J_{W,i}^{abs}(\mathbf{X}, {}^{\mathcal{W}\!}\mathbf{X}) = \left(1 - \lambda_2 \cdot \xi({}^{\mathcal{W}\!}\mathbf{k}_i, \mathbf{k}_i)\right)^2, \quad 
\xi(\mathbf{y}, \mathbf{z}) = \frac{\mathbf{y} \cdot \mathbf{z}}{\|\mathbf{y}\|\|\mathbf{z}\|}
\label{eq:J_W_single}
\end{equation}

\textbf{Line of Sight Cost (\(J_S\))}: Penalizes deviations from line-of-sight vectors connecting the camera's focal point and BlazePose's normalized landmarks.

\begin{equation}
{J}_{S}(\mathbf{X}, {}^{\mathcal{N}\!}\mathbf{X}) = \frac{1}{K} \sum_{i=1}^{K} J_{S,i} = \frac{1}{K} \sum_{i=1}^{K}\left(1 - \lambda_3 f(v_i) \cdot \xi(\mathbf{x}_i, \mathbf{v}_i)\right)^2
\label{eq:search_vec_costs}
\end{equation}

with \(K\) representing the total number of landmarks. \(\mathbf{x}_i\) is the position of the \(i\)-th optimized landmark and \(\mathbf{v}_i\) is the line of sight vector of the same landmark from the normalized landmarks \({}^{\mathcal{N}\!}\mathbf{X}\).

The visibility weighting function \(f(v_i)\) is used to put more trust on landmarks with a high visibility score \(v_i\):

\begin{equation}
f(v_i) = \frac{1}{2} + \frac{1}{2} e^{-\lambda_4 (1 - v_i)}
\label{eq:adjustment_factor_visibility}
\end{equation}

Parameters \(\lambda_3, \lambda_4\) adjust the visibility weighting and scaling, respectively.

\textbf{Bone Ratio Costs (\(J_B, J_M\))}: These enforce anatomical plausibility by penalizing deviations from expected bone ratios. The individual bone ratio cost (\(J_B\)) for bones (ulna, humerus, pelvis, femur, tibia) is calculated as:

\begin{equation}
{J}_{B}(\mathbf{X}, \mathbf{B}) = \frac{1}{R} \sum_{i=1}^{R} \left(\lambda_5 (b_i - b_i^*)\right)^2
\label{eq:ratio_cost}
\end{equation}

where \(b_i\) are the bone ratios of the optimized skeleton computed from \(\mathbf{X}\) and \(b_i^*\) are the estimated bone ratios as result of the bone ratio estimation \ref{subsec:boneratioadjustment}. \({R}\) is the number of rigid bones modeled in the skeleton.

The multi-bone ratio cost (\(J_M\)) for segments affected by shoulder motion (shoulder width, lateral trunk, spine) incorporates humerothoracic elevation angles \(\beta_L,\beta_R\):

\begin{equation}
{J}_{M}(\mathbf{X}, \mathbf{B}) = \frac{1}{M} \sum_{i=1}^{M}\left(m_i - u_i(m_i^*, \beta_L, \beta_R)\right)^2
\label{eq:total_ratio_cost_m}
\end{equation}

Here, \(m_i\) is the predicted bone ratio, \(m_i^*\) is its expected ratio. \(u_i\) are the polynomial functions which adjust the ratios according to elevation angles as defined in Section \ref{subsec:humerothoracic-elevation-functions}. \(M\) represent the number of  multi-bone segments. To stabilize optimization, if bone ratios remain consistent across frames, the weights for \(J_M\) and \(J_B\) are increased, strengthening their influence in the cost function.

\subsubsection{Optimization}
\label{subsec:optimization}

To obtain optimized landmark positions, the total cost is minimized. The initial guess for each frame is the optimized result of the previous frame, which improves convergence speed and promotes tracking of consistent local minima compared to using a standard position as an initial guess. For the first frame, the BlazePose World coordinates are used for initialization. The L-BFGS-B optimization algorithm is used to efficiently minimize the cost function. We use the implementation from \texttt{scipy.optimize.minimize} \cite{virtanen2020scipy}.

\section{Experiments}
\subsection{Physio2.2M Dataset}
The Physio2.2M dataset is an Akina internal dataset; access can be requested by contacting the authors of the article published in Nature Scientific Reports \cite{fastape2024}. It was collected using a Vicon motion capture system with 27 infrared cameras (200 Hz) and synchronized RGB video (30 Hz). It consists of 25 participants (11 male, 14 female, aged 20–33) performing seven exercises: squats, bridge, bird, abduction, shoulder, knee, and stretch, each repeated five times.

The dataset emphasizes challenging poses common in physiotherapy, including supine and quadruped positions, providing a benchmark for evaluating markerless pose estimators. Ground truth joint positions were obtained from 46 anatomical markers. 

\begin{table}[htbt]
\centering
\begin{tabular}{lccc}
\hline
\textbf{Group} & \textbf{Age (years)} & \textbf{Height (mm)} & \textbf{Weight (kg)} \\ \hline
Male (n=11)    & 26 (23--31)          & 1818 (1730--1920)    & 79 (65--94)          \\
Female (n=14)  & 26 (20--33)          & 1649 (1580--1740)    & 60 (47--80)          \\
Overall (n=25) & 26 (20--33)          & 1724 (1580--1920)    & 68 (47--94)          \\ \hline
\end{tabular}
\caption{Participant demographics in the Fast Ape dataset.}
\label{tab:fastape-participants}
\end{table}

A total of 321 videos (346,297 frames) were used for evaluation.

\subsection{Assessment Methodology}
\label{subsec:assessment_methodology}

To compare the estimated and ground truth poses, all landmark skeletons have the origin located at the pelvis and are scaled using an optimizer to minimize position differences. This alignment ensures a fair absolute position comparison for estimated poses in different units.

\textbf{Mean Per Joint Position Error (MPJPE)} is used to assess landmark accuracy:
\begin{equation}
\overline{MPJPE} = \frac{1}{N_{frames}} \sum_{i=1}^{N_{frames}} \frac{1}{N_{joints}} \sum_{j=1}^{N_{joints}} PE_j^i
\label{eq:average_mpjpe}
\end{equation}
where \( PE_j^i \) is the Euclidean distance between the estimated and ground truth coordinates for joint \(j\) in frame \(i\).

\textbf{Angle Accuracy} is evaluated for key body angles (elbow, shoulder, knee, hip flexion), calculated from joint positions. The angle error is defined as:
\begin{equation}
AE_j = \frac{1}{N_{frames}} \sum_{i=1}^{N_{frames}}\left| \theta_j^i - {}^{\mathcal{G}\!}\theta_j^i \right|
\label{eq:average_angle_error_per_angle}
\end{equation}
where \( \theta_j^i \) is the estimated angle and \( {}^{\mathcal{G}\!}\theta_j^i \) the ground truth.

\section{Results}
\subsection{Initial Bone Ratios}
\label{sec:res_initial_bone_ratios}

The initial bone ratios obtained with the methodology outlined in Section \ref{subsec:initial_bone_ratios}, along with the average bone ratios from the ANSUR2 dataset \cite{ansur2}, and the final ratios derived after optimization on the Physio2.2M dataset, are presented in Table~\ref{tab:bone_ratios_comparison}.

\begin{table}[htbt]
\centering
\begin{tabular}{l r c r}
\toprule
\textbf{Segment} & \textbf{Initial} & \textbf{ANSUR2} & \textbf{Final} \\
\midrule
Ulna & 0.0862 & 0.0853 & 0.0932 \\
Humerus & 0.1015 & 0.1077 & 0.0976 \\
Femur & 0.1462 & 0.1341 & 0.1430 \\
Tibia & 0.1297 & 0.1278 & 0.1389 \\
Pelvis & 0.0728 & 0.0903 & 0.0690 \\
Shoulder Width & 0.1260 & - & 0.1100 \\
Spine & 0.1859 & - & 0.1647 \\
Lateral Trunk & 0.1878 & - & 0.1669 \\
\bottomrule
\end{tabular}
\caption{Comparison of Initial, ANSUR2, and Final bone ratios.}
\label{tab:bone_ratios_comparison}
\end{table}

\subsection{MPJPE Results}
\begin{figure}[htbt]
    \centering
    \includegraphics[width=\textwidth]{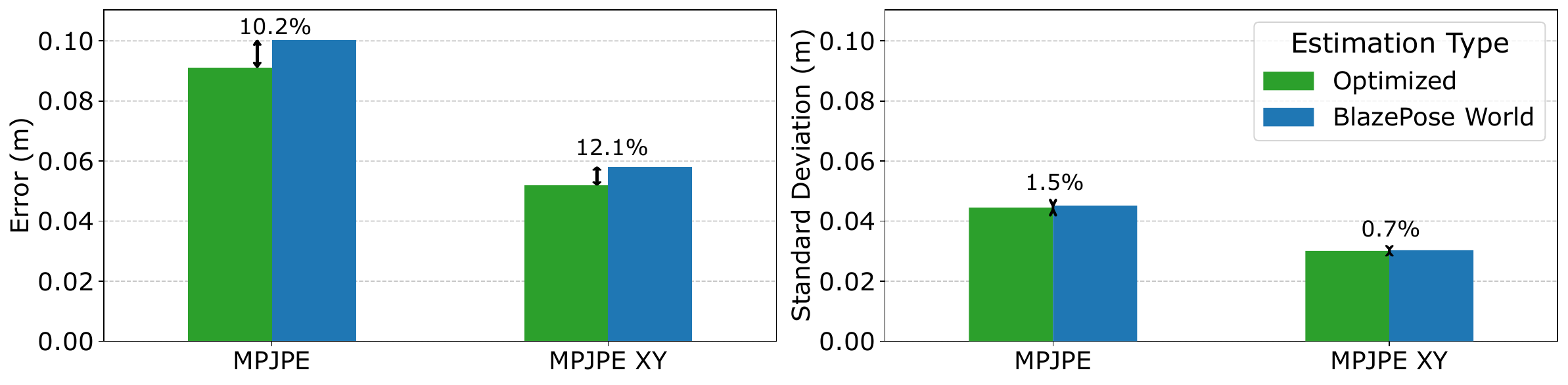}
    \caption{The left plot shows the average MPJPE in 3D and the XY plane (accuracy). The right plot displays the standard deviation of MPJPE (precision).}
    \label{fig:fastape_overall_MPJE_error_and_stdv}
\end{figure}

Figure~\ref{fig:fastape_overall_MPJE_error_and_stdv} illustrates the overall MPJPE results. The left plot shows the average MPJPE in 3D and the XY plane, reflecting the accuracy of the posture estimation. On the right, the standard deviation of the MPJPE for these dimensions is displayed, providing insights into the precision of the predictions.

\subsection{Body Angle Results}
\begin{figure}[htbt]
    \centering
    \includegraphics[width=\textwidth]{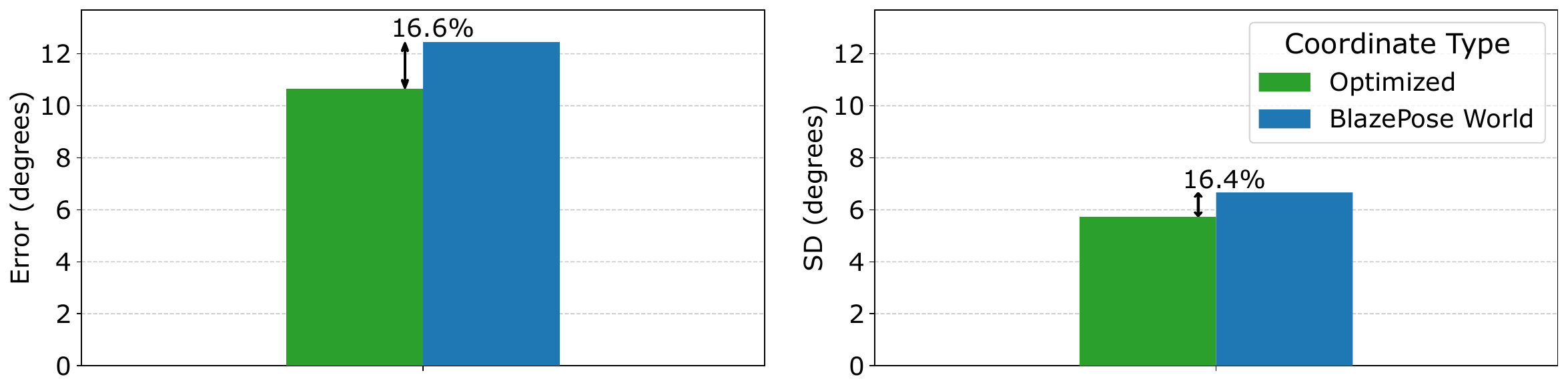}
    \caption{Comparison of average angle error and standard deviation for all body angles.}
    \label{fig:fastape_overall_body_angles_comparision}
\end{figure}

Figure~\ref{fig:fastape_overall_body_angles_comparision} summarizes the overall error and standard deviation for all body angles, comparing the optimized and BlazePose world coordinates.

\begin{figure}[htbt]
    \centering
    \includegraphics[width=\textwidth]{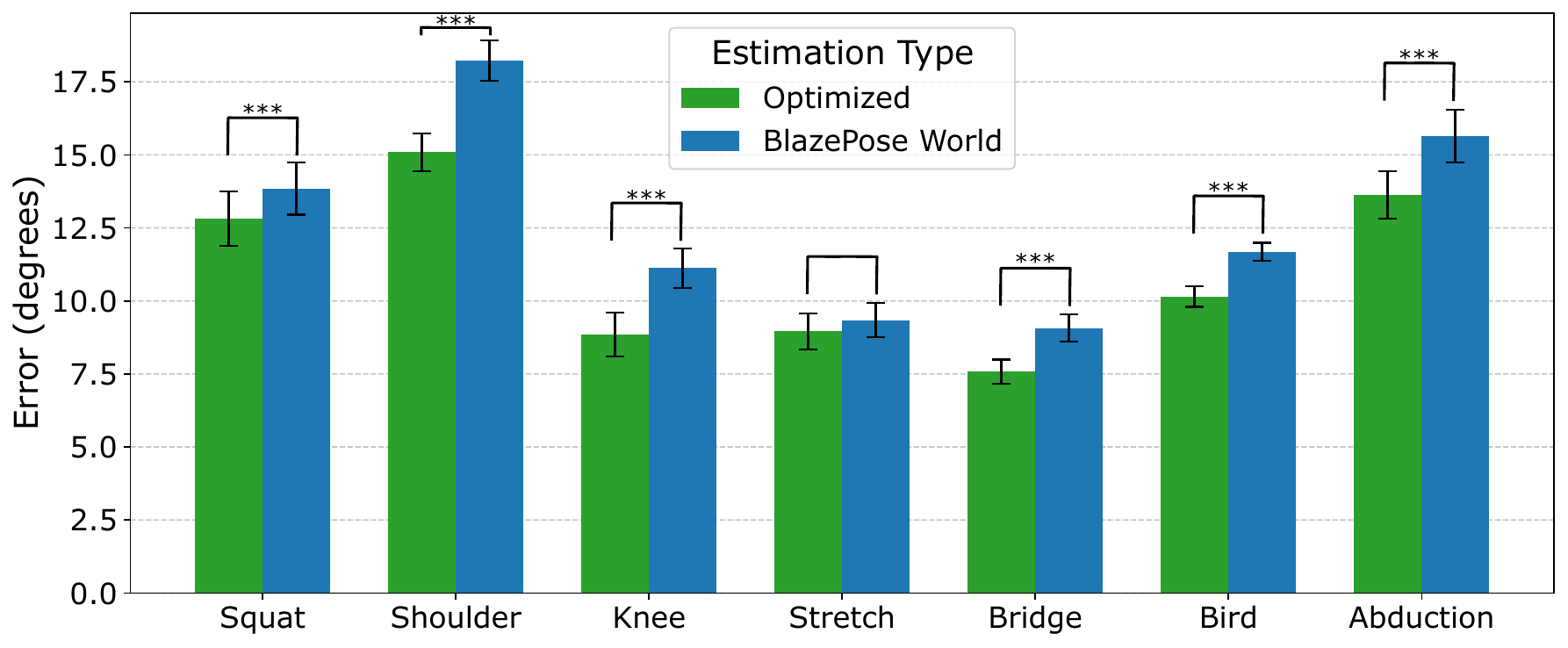}
    \caption{Average angle error by exercise.}
    \label{fig:fastape_overall_body_angles_comparision_ex_wise}
\end{figure}

Figure~\ref{fig:fastape_overall_body_angles_comparision_ex_wise} presents the average angle error grouped by exercise type, illustrating variations in performance across different movements.

\subsection{Statistical Significance of Results}
Table~\ref{tab:body_angles_p_values} and Table~\ref{tab:p_values_exercises_updated} present the p-values for body angles and landmarks across exercises, with the "Overall" column aggregating results across all exercises. Abbreviations are used for readability, where L denotes left and R denotes right:

\textbf{Body Angles (BA):}  
LEF / REF (Elbow Flexion), LHF / RHF (Hip Flexion), LKF / RKF (Knee Flexion), LSE / RSE (Shoulder Elevation).  

\textbf{Landmarks (LM):}  
LA / RA (Ankle), LE / RE (Elbow), LH / RH (Hip), LK / RK (Knee), LS / RS (Shoulder), LW / RW (Wrist).  

\begin{table}[htbt]
\centering
\begin{tabular}{@{}lcccccccc@{}}
\toprule
\textbf{LM}          & \textbf{Squat} & \textbf{Shoulder} & \textbf{Abduction} & \textbf{Bird} & \textbf{Bridge} & \textbf{Stretch} & \textbf{Knee} & \textbf{Overall} \\ \midrule
\textbf{LA}       & 0.4275 & 0.0520 & 0.0077 & 0.0000 & 0.0000 & 0.0000 & 1.0000 & 0.0000 \\
\textbf{LE}       & 0.0007 & 0.0000 & 0.0000 & 0.2475 & 0.0003 & 0.4428 & 0.0842 & 0.0000 \\
\textbf{LH}         & 0.0000 & 0.0000 & 0.0000 & 0.0000 & 0.0000 & 0.0000 & 0.0000 & 0.0000 \\
\textbf{LK}        & 0.1589 & 1.0000 & 1.0000 & 0.0000 & 0.0000 & 0.0000 & 0.0009 & 0.0000 \\
\textbf{LS}    & 0.0000 & 0.0000 & 0.0026 & 1.0000 & 0.3100 & 1.0000 & 0.0000 & 0.0000 \\
\textbf{LW}       & 0.0000 & 0.0000 & 0.0000 & 1.0000 & 1.0000 & 1.0000 & 0.2836 & 1.0000 \\
\textbf{RA}      & 1.0000 & 0.0478 & 0.0003 & 1.0000 & 0.0026 & 0.0003 & 0.0030 & 0.0000 \\
\textbf{RE}      & 0.0027 & 0.0000 & 0.0001 & 0.0000 & 0.0105 & 1.0000 & 0.0000 & 0.0000 \\
\textbf{RH}        & 0.0000 & 0.0000 & 0.0000 & 0.0000 & 0.0000 & 0.0000 & 0.0000 & 0.0000 \\
\textbf{RK}       & 1.0000 & 1.0000 & 1.0000 & 0.0000 & 0.0000 & 0.0000 & 0.0000 & 0.0000 \\
\textbf{RS}   & 0.0000 & 0.0000 & 0.0000 & 0.0000 & 0.0000 & 0.2588 & 0.0000 & 0.0000 \\
\textbf{RW}      & 0.0002 & 0.0000 & 0.0000 & 1.0000 & 1.0000 & 1.0000 & 0.0000 & 0.0000 \\ \bottomrule
\end{tabular}
\caption{P-values for all landmarks and exercises.}
\label{tab:p_values_exercises_updated}
\end{table}

\begin{table}[htbt]
\centering
\begin{tabular}{@{}lcccccccc@{}}
\toprule
\textbf{BA} & \textbf{Squat} & \textbf{Shoulder} & \textbf{Abduction} & \textbf{Bird} & \textbf{Bridge} & \textbf{Stretch} & \textbf{Knee} & \textbf{Overall} \\ \midrule
\textbf{LEF}   & 1.0000 & 0.0000 & 0.0000 & 0.4102 & 1.0000 & 0.0222 & 1.0000 & 0.0000 \\
\textbf{LHF}   & 0.0000 & 0.0000 & 0.0000 & 0.0491 & 0.0002 & 0.4517 & 0.0000 & 0.0000 \\
\textbf{LKF}   & 0.0000 & 0.0000 & 0.0000 & 0.0000 & 0.0000 & 1.0000 & 0.0000 & 0.0000 \\
\textbf{LSE}   & 1.0000 & 0.0000 & 0.0000 & 1.0000 & 0.0654 & 0.0397 & 0.0217 & 0.0000 \\
\textbf{REF}   & 1.0000 & 0.0000 & 0.0000 & 0.0000 & 0.0000 & 0.0002 & 0.0000 & 0.0000 \\
\textbf{RHF}   & 0.0000 & 0.0001 & 0.0000 & 0.0886 & 0.0802 & 1.0000 & 0.0000 & 0.0000 \\
\textbf{RKF}   & 0.0000 & 0.0000 & 0.0001 & 0.0000 & 0.0000 & 0.1013 & 0.0000 & 0.0000 \\
\textbf{RSE}   & 0.0000 & 0.0000 & 0.0000 & 1.0000 & 0.0000 & 1.0000 & 1.0000 & 0.0000 \\ 
\bottomrule
\end{tabular}
\caption{P-values for all body angles and exercises.}
\label{tab:body_angles_p_values}
\end{table}

\subsection{Ablation Study}
\label{sec:ablation_study}

Table~\ref{tab:ablation_study_metric_comparison} presents a comparison of key metrics across three configurations: BlazePose world coordinates, optimized coordinates with bone length ratios reset at the start of each video, and optimized coordinates where bone ratios from the previous video were reused for the same participant. The metrics include 3D mean per joint position errors (MPJPE), their standard deviations, and average angle errors along with standard deviations.

\begin{table}[htbt]
\centering
\begin{tabular}{|l|c|c|c|}
\hline
\textbf{Metric} & \textbf{BP World} & \textbf{Optimized (Reset)} & \textbf{Optimized (Reused)} \\ \hline
3D MPJPE & 100.3mm & 91.6mm & 91.0mm \\ \hline
3D Std Dev & 45.2mm & 44.6mm & 44.5mm \\ \hline
Avg Angle Error & 12.3° & 10.9° & 10.6° \\ \hline
Avg Angle Std Dev & 6.6° & 5.8° & 5.7° \\ \hline
\end{tabular}
\caption{Comparison of metrics between BlazePose world coordinates, optimized coordinates with bone ratios reset at the start of each video, and optimized coordinates with reused bone ratios for the same participant.}
\label{tab:ablation_study_metric_comparison}
\end{table}

\subsection{Bone Ratio Results}

\begin{figure}[htbt]
    \centering
    \includegraphics[width=\textwidth]{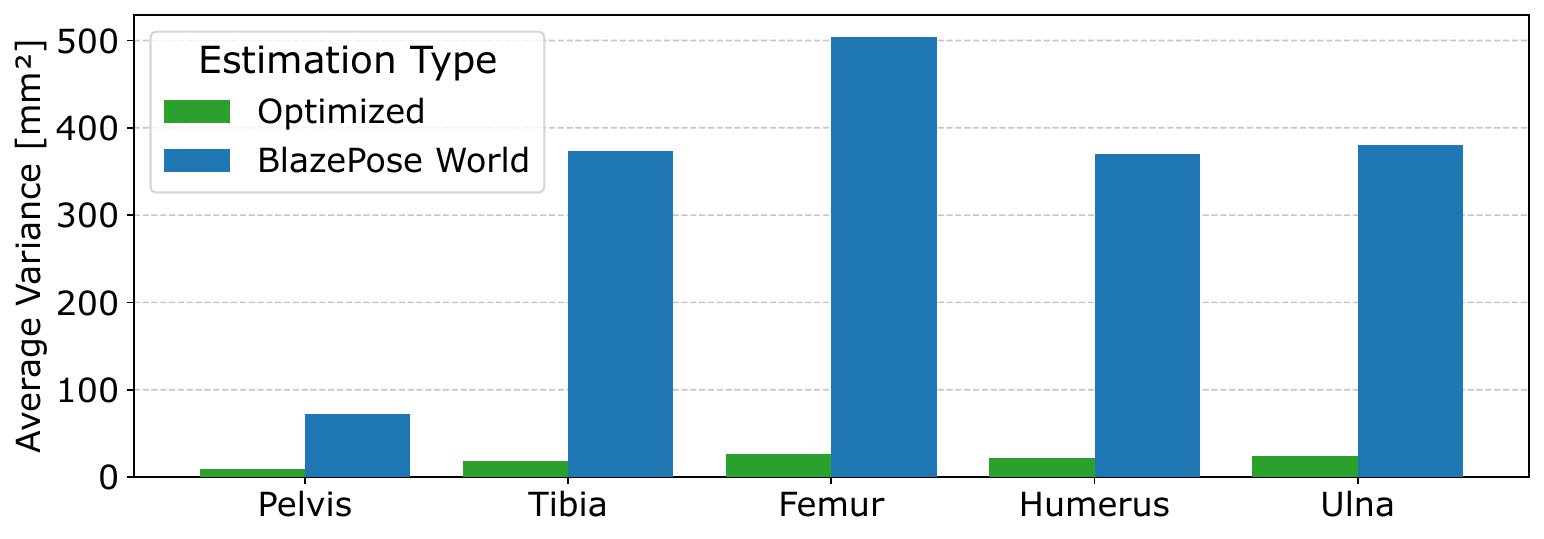}
    \caption{Variance Comparision of Optimized and BlazePose World Coordinates for all bones.}
    \label{fig:fastape_bone_variance_comparision}
\end{figure}

Figure \ref{fig:fastape_bone_variance_comparision} shows the average variance for each bone in BlazePose World and Optimized coordinates.

On average the bone length variance of BlazePose World coordinates is \(370\text{mm}^2\) while for the Optimized Coordinates the average bone length variance is \(21\text{mm}^2\).

\section{Discussion}
\subsection{Key Findings and Contributions}

Our optimization pipeline demonstrates significant improvements in pose estimation accuracy across various metrics. On the Physio2.2M dataset, our pipeline reduced the 3D MPJPE by 10.2\%, and the MPJPE in the XY plane by 12.1\% (Figure \ref{fig:fastape_overall_MPJE_error_and_stdv}). Our pipeline improved body angles by 16.6\%, achieving consistent gains across all exercises (Figure \ref{fig:fastape_overall_body_angles_comparision_ex_wise}). These results highlight the robustness of the proposed optimization framework in real-world scenarios.

The optimization framework maintains consistent bone lengths across frames. This is evident in the significantly reduced bone variance compared to BlazePose world coordinates (Figure \ref{fig:fastape_bone_variance_comparision} average bone length variance is reduced by 94.3\%).

\subsection{Limitations of the Experiments}
The Physio2.2M dataset, while representing the state-of-the-art in motion capture systems and providing an extensive evaluation basis with over 300'000 frames, has certain limitations. The motion capture system fixes only the hip-to-hip distance, while other bone lengths can vary by up to 2 cm \cite{Baker2018}. In rare cases, we observed limb length variations of 25\%, likely caused by deficient placement of markers or displacement due to soft tissue movement.

\subsection{Limitations of the Methods}
As shown in Table \ref{tab:bone_ratios_comparison}, the average final bone ratios differ noticeably from the initial ratios. Specifically, the tibia and ulna become longer, while the spine and lateral trunk appear shorter. Ideally, these ratios should be more consistent. The underlying issues stem from limitations in BlazePose estimations: the tibia and ulna are often inaccurately estimated as not lying completely in the \(xy\)-plane, even when they do in reality, resulting in overestimated lengths. Conversely, the spine encounters the opposite problem. When participants lean forward or backward, BlazePose frequently underestimates the tilting angle, leading to shorter spine estimates. 

Addressing these discrepancies is challenging, as it requires distinguishing between incorrect BlazePose estimations and individuals whose bone ratios naturally deviate from anthropometric norms.

\subsection{Importance for the Field}
This work significantly advances monocular pose estimation by addressing critical limitations of existing models like BlazePose, HRNet, and OpenPose, which fail to incorporate anthropometric and biomechanical constraints. We demonstrated that the inclusion of anatomical constraints and biomechanical knowledge can enhance the posture estimation's accuracy and achieve consistent bone lengths across frames. Our pipeline achieves a 10.2\% improvement in the widely used MPJPE metric compared to state-of-the-art models, alongside a 16.6\% reduction in angle errors. 

Moreover, the proposed method operates as a post-processing step without direct access to image data, relying solely on landmark coordinates provided by existing pose estimation frameworks. Consequently, it can be executed entirely on backend servers, ensuring user data privacy and avoiding additional computational load on end-user hardware.

These advancements are relevant for real-world applications of action recognition and applications requiring precise biomechanical analysis, such as such as clinical gait assessments, sports coaching, or remote physiotherapy.

\subsection{Outlook}
While the proposed optimization pipeline shows significant improvements in monocular pose estimation accuracy, several challenges remain that provide avenues for future research.

Occlusion handling remains a critical challenge, particularly for distal joints such as the wrist, where errors propagate due to missing or occluded landmarks. Future research could focus on the development of occlusion-aware models that infer missing joint positions by leveraging correlations of visible landmarks and knowledge of biomechanical dynamics.

To address the issue that the final bone ratios differ significantly from the initial ones, a potential solution could be to create a new dataset and derive a model from it. This model would then be specifically designed to estimate the true lengths of problematic bones. By accurately determining the actual bone lengths, one could analyze the average deviations BlazePose produces in estimations across various angles relative to the \(xy\)-plane. This would enable the implementation of correction factors that adjust for these systematic errors, effectively mitigating a significant portion of the issue and improving the reliability of bone ratio estimations.

\section{Conclusion}
This work introduces a novel optimization pipeline for human pose estimation with monocular videos, designed to address key limitations of state-of-the-art models like BlazePose. By integrating anthropometric constraints, biomechanical knowledge, and advanced filtering techniques, the proposed framework achieves substantial improvements in accuracy, consistency, and robustness, as evidenced by evaluations on a diverse dataset.

The pipeline attained an average body angle accuracy of 10.6°, exceeding the target of 12° considered a requirement for motion coaching. Additionally, it ensured close to consistent bone lengths throughout movements, addressing a critical limitation in pose estimation models like BlazePose.

Compared to BlazePose, our pipeline achieves a remarkable reduction of 10.2\% in 3D MPJPE and a 12.1\% reduction in MPJPE in the XY plane on the Physio2.2M dataset. These improvements are complemented by a 16.6\% reduction in body angle errors, underscoring the pipeline's ability to estimate both joint positions and body angles with substantially higher accuracy compared to related work.

Bone length variation could be reduced by 94.3\% compared to BlazePose. This consistency is especially vital in applications such as physiotherapy and sports analysis, where reliability and anatomical plausibility are paramount. 

This was accomplished by post-processing of landmark data only. Thus, the computational load of the added pipeline components can be managed by a backend server while the videos containing sensitive data are not required on the backend. Thereby, dataprivacy can be maintained without increasing the computational demand on the user's hardware.

In conclusion, the presented method achieves a significant improvement over the BlazePose model. The achieved human posture estimation performance should facilitate new features in automated coaching, e.g., in health care applications.

\par\vfill\par
\clearpage
\bibliographystyle{splncs04}
\bibliography{main}
\end{document}